\documentclass[runningheads]{llncs}

\usepackage{eccv}

\usepackage{eccvabbrv}

\usepackage{graphicx}
\usepackage{booktabs}
\usepackage{amsmath,amssymb}
\usepackage{multirow}
\usepackage{xcolor}

\usepackage{makecell}
\usepackage{siunitx}
\usepackage{textcomp}

\usepackage{tabularx}

\usepackage{placeins}
\usepackage{xspace}

\usepackage{pifont}
\newcommand{\cmark}{\ding{51}} % ✓
\newcommand{\xmark}{\ding{55}} % ✗

\usepackage[accsupp]{axessibility}

\usepackage[breaklinks,colorlinks,citecolor=eccvblue]{hyperref}
\usepackage[capitalize]{cleveref}

\begin{document}

% ---------------------------------------------------------------
% Title / running head
\title{CLIP-Guided SAM: Parameter-Efficient Semantic Conditioning for Promptable Segmentation}
% CLIP-Guided SAM: Parameter-Efficient Semantic Conditioning for Promptable Segmentation
% CLIP-Guided SAM: A Parameter-Efficient Fine-Tuning Framework for Prompt-Driven Semantic Segmentation
\titlerunning{CLIP-Guided SAM}

% ---------------------------------------------------------------
% Authors
\author{Shayan Jalilian \and Abdul Bais}
\authorrunning{S. Jalilian and A. Bais}
\institute{University of Regina, Regina, SK, Canada \\
\email{sjs949@uregina.ca, Abdul.Bais@uregina.ca}}

\maketitle

% ---------------------------------------------------------------
% Abstract (was thesis \section{Abstract})
\begin{abstract}

Promptable foundation models such as the Segment Anything Model (SAM) produce high-quality masks but remain semantically blind, relying on external prompts to specify categories. Existing vision--language approaches address this limitation by using external prompt coupling, in which a vision--language model generates spatial prompts for SAM as a separate stage.

We propose \textbf{CLIP-Guided SAM}, a parameter-efficient segmentation framework built on {internal semantic conditioning}. Instead of using semantic signals only to generate prompts, we inject CLIP-derived text, vision, and similarity features directly into SAM’s image encoder via lightweight multi-modal semantic adapters. These adapters condition SAM’s internal feature representations, allowing semantic information to influence mask prediction while preserving SAM’s original promptable interface.

Our framework is designed for low labelled-data settings and applies to both general-domain benchmarks and specialized downstream tasks. It supports two operating modes: {Manual mode}, for interactive segmentation with both text and spatial prompts, and {Semi-Automatic (text-only) mode}, for applications that require concept-specific segmentation using only textual input. We show that robustness depends on aligning training with the type of prompts used at inference, making train--test prompt consistency an important design principle.

Through extensive experiments and ablations, we evaluate our method against SAM+PEFT baselines without semantic conditioning, vision--language + SAM pipelines, SAM~3, and strong semi-supervised segmentation methods that rely on large amounts of unlabelled data. Across these settings, CLIP-Guided SAM consistently achieves superior or competitive performance while remaining parameter-efficient in both training and deployment.

\end{abstract}

% ---------------------------------------------------------------
% Main paper content (was thesis \input's after \chapter)
% IMPORTANT: Ensure these section files start with \section{...}, not \chapter{...}

\section{Introduction}
\label{sec:intro}

Prompt-based segmentation, popularized by the Segment Anything Model (SAM)\allowbreak~\cite{kirillov2023segment}, has significantly expanded the flexibility of visual localization. With minimal spatial input—such as a point, box, or mask—SAM can produce high-quality object masks across diverse scenes. However, SAM is intentionally \emph{class-agnostic}: it excels at grouping visual structures but does not encode semantic category information. As a result, identifying \emph{what} to segment remains an external problem. Furthermore, SAM is sometimes adapted via Parameter-Efficient Fine-Tuning (PEFT) methods for specific tasks to enhance performance and address label scarcity or computational constraints.

Vision–language models (VLMs) such as Contrastive Language–Image Pre-training (CLIP)~\cite{radford2021clip} provide complementary strengths, offering semantic representations aligned with natural language but lacking the spatial precision required for dense segmentation. This complementarity has motivated VLM+SAM systems based on \emph{external prompt coupling}, where a VLM generates spatial prompts—points, boxes, or masks—that are provided to a frozen or lightly adapted SAM.

\begin{figure}[t]
    \centering
    \includegraphics[width=\textwidth]{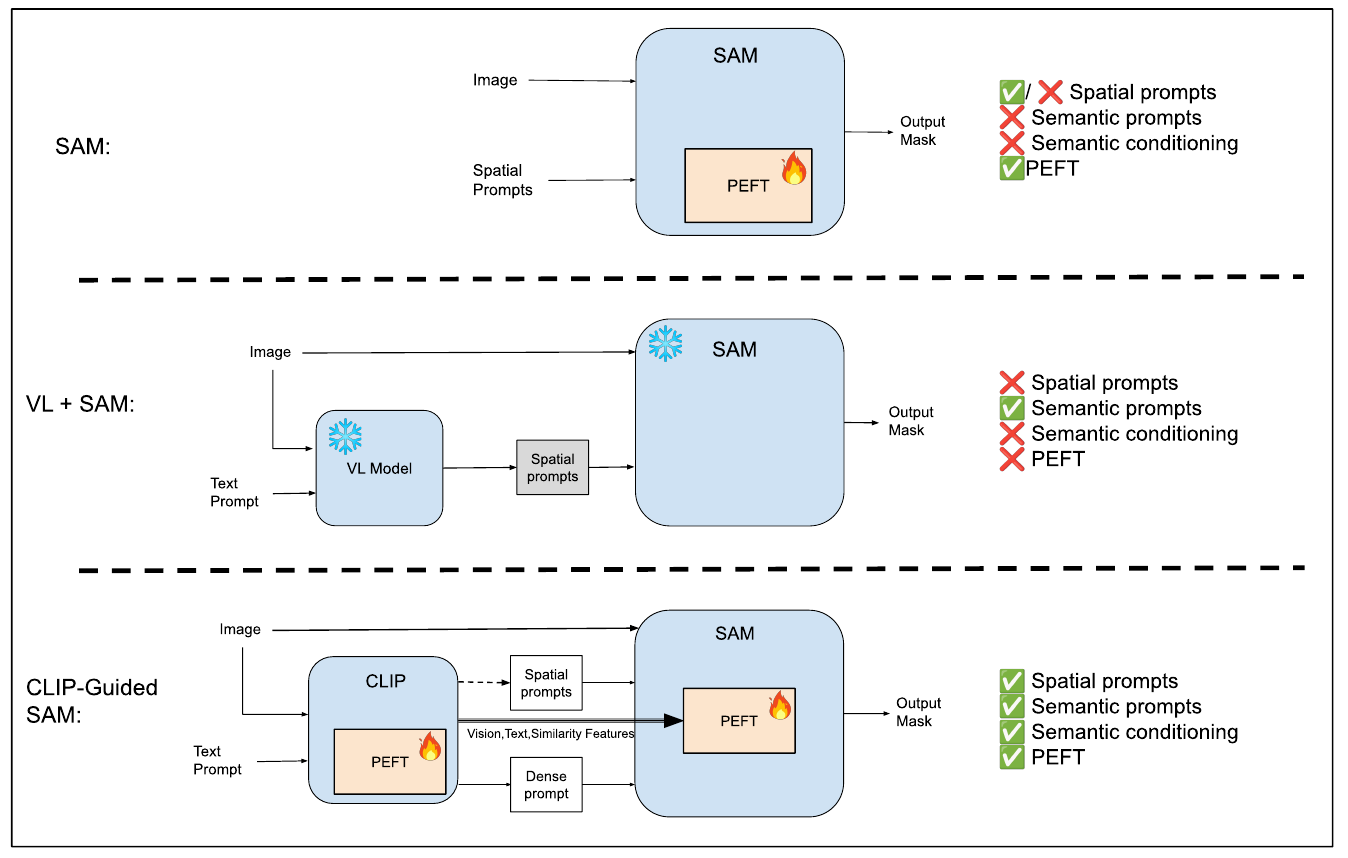}
    \vspace{-0.5em}
    \caption{
    \textbf{CLIP-Guided SAM Overview.}
    Comparison of (top) \emph{SAM with PEFT}, (middle) \emph{typical VLM+SAM pipelines}, and (bottom) our \emph{CLIP-Guided SAM}.
    Existing approaches either rely on spatial prompting alone or couple a VLM to SAM externally via prompt generation, leaving SAM’s internal representations semantically uninformed.
    Our approach introduces \textbf{internal semantic conditioning} by injecting CLIP-derived features directly into SAM’s image encoder through multi-modal adapters, enabling joint SAM--CLIP co-adaptation.
    }
    \vspace{-1em}
    \label{fig:teaser}
\end{figure}

While effective in some settings, this stage-wise and shallow integration keeps semantic reasoning and spatial segmentation architecturally separate. Semantic information influences SAM only indirectly through sparse or noisy prompts, leaving its internal feature representations semantically uninformed. This limitation is even more pronounced in low-label regimes and text-only scenarios, where supervision is limited and spatial cues are suboptimal. \Cref{fig:teaser} illustrates the difference between existing approaches—such as SAM with PEFT or training-free VLM+SAM pipelines—and our proposed framework, which injects semantic signals directly into SAM’s internal representations.

In this work, we propose \emph{CLIP-Guided SAM}, a parameter-efficient framework that unifies \textbf{internal semantic conditioning} with \textbf{external spatial prompting}. Rather than relying solely on prompt-level guidance, we inject CLIP-derived text, vision, and similarity features directly into SAM’s image encoder via semantic adapters. This design allows semantic information to influence feature formation during segmentation, while spatial prompts remain complementary for boundary refinement and localization.

A central component of our framework is the \textbf{joint co-adaptation} of SAM and CLIP. Through systematic ablations, we show that effective semantic conditioning requires optimizing CLIP not merely as a prompt generator, but as an internal semantic guide tailored to SAM’s feature space. Freezing CLIP or pre-fine-tuning it independently beforehand to improve spatial prompt quality leads to inferior performance, whereas end-to-end co-adapting both models enables CLIP to learn to produce semantic signals useful for segmentation and better aligned with SAM's feature space. We further observe asymmetric sensitivity in tuning budgets: performance saturates quickly with additional SAM backbone tuning, whereas increasing the number of trainable parameters in CLIP produces larger gains.

CLIP-Guided SAM is designed to support multiple deployment scenarios. In an \emph{interactive} mode, the framework combines text prompts with user-provided point inputs, modelling user-in-the-loop segmentation or annotation workflows. In a \emph{text-only} (semi-automatic) mode, segmentation is driven solely by class descriptions, without human spatial input. In this setting, spatial prompts serve as model-generated inputs rather than supervision. We therefore introduce a \textbf{train-test-prompt-aligned training} strategy that trains directly on noisy prompts sampled from CLIP similarity maps. This alignment mitigates the train–test mismatch introduced by idealized ground-truth clicks and yields substantially more robust inference under text-only prompting.

% We evaluate CLIP-Guided SAM across both general semantic segmentation benchmarks (COCO, ADE20K, PASCAL VOC) and specialized downstream tasks such as camouflaged object detection. Under limited supervision and closed-set assumptions, our method consistently outperforms SAM-based parameter-efficient baselines and achieves competitive performance relative to heavier vision–language systems. In particular, with a few hundred labelled images, our modular co-adaptation framework approaches the zero-shot performance of recent large-scale models such as SAM 3, while using significantly fewer parameters and enabling efficient tuning on commodity GPUs. These results highlight a practical trade-off between large, unified models optimized for zero-shot generalization and modular, parameter-efficient systems tailored for domain adaptation under label, memory, and computational constraints.

We evaluate CLIP-Guided SAM on general semantic segmentation benchmarks (COCO, ADE20K, PASCAL VOC) and specialized downstream tasks such as camouflaged object detection. Under limited supervision, our method consistently outperforms SAM-based parameter-efficient baselines and achieves competitive performance relative to heavier vision–language systems. With only a few hundred labelled images, our modular co-adaptation framework approaches the zero-shot performance of large-scale models such as SAM~3, while using substantially fewer parameters and enabling efficient tuning on commodity GPUs.

\noindent
\textbf{Contributions.} In summary, we make the following contributions:
\begin{itemize}
    \item We introduce a framework for \textbf{internal semantic conditioning} of SAM, shifting VLM+SAM integration from external prompt coupling to joint \allowbreak feature-level co-adaptation.
    \item We analyze \textbf{co-adaptation dynamics} between SAM and CLIP, showing that adapting the semantic guide is critical, whereas additional tuning of the SAM backbone yields diminishing returns.
    \item We propose a \textbf{train-test prompt alignment} strategy for text-only segmentation that improves robustness by matching training and inference prompt distributions.
    % \item We present a framework applicable to both general and specialized segmentation tasks. We also provide two design forms, suited to two distinct applications and philosophies: interactive segmentation (human-provided point and text prompts, manual mode) and text-prompted segmentation (text-only mode). Our framework is designed for low-label and modest-compute settings.
    \item We present a parameter-efficient system applicable to both general and specialized segmentation tasks, supporting interactive (text + spatial prompts) and text-only deployment modes under low-label and modest-compute settings.
    \item We demonstrate strong performance on different comparisons and experiments, along with comprehensive ablations, highlighting an efficient and strong alternative to SAM-PEFT baselines, VLM+SAM pipelines, large foundation models, and even strong semi-supervised methods in certain contexts.
        
    \item We demonstrate strong empirical performance across diverse comparisons and ablations, providing an efficient and competitive alternative to SAM-PEFT baselines, VLM+SAM pipelines, large segmentation foundation models, and, in certain regimes, semi-supervised methods.

\end{itemize}

\section{Related Works}

\subsection{SAM and Parameter-Efficient Fine-Tuning}

SAM is a powerful prompt-based segmentation foundation model, but often requires adaptation for specialized downstream tasks \cite{chen2023samfails}. Because full fine-tuning is costly in low-label regimes, parameter-efficient fine-tuning (PEFT) methods such as adapters \cite{houlsby2019parameter}, LoRA \cite{hu2022lora}, and prompt tuning \cite{lester2021prompttuning} are widely used.

Prior work adapts SAM via lightweight modules or modified heads, e.g., SAM-Adapter \cite{chen2023SamAdptr}, Conv-Meets-LoRA \cite{zhong2024convMLoRA}, SU-SAM \cite{song2024SUSAM}, and TS-SAM \cite{yu2024tssam}, while others provide light semantic conditioning through text embeddings, e.g., SAM-PTx \cite{jalilian2025samptx}. In contrast, we keep SAM’s prompt encoder and mask decoder intact and inject CLIP-derived text, vision, and similarity features into the image encoder via semantic adapters, enabling text-prompted semantic conditioning with minimal architectural change.

\subsection{Vision--Language Models for Semantic Segmentation}

Vision–language models such as CLIP \cite{radford2021clip} align images and text in a shared embedding space and have been widely used for segmentation. Zero-shot and weakly supervised approaches derive activation maps or grouped regions from CLIP representations \cite{cha2023learning, shin2022reco, xu2022groupvit, zhou2022maskclip, cho2024cat, gu2021open, luo2023segclip}, but often suffer from coarse boundaries due to limited spatial precision \cite{hoyer2024semivl}. 

Open-vocabulary segmentation methods instead train large models with language supervision to generalize to unseen classes \cite{ghiasi2022scaling, li2022language, ding2022decoupling, xu2022simple, liang2023open, zhou2023zegclip}. SemiVL \cite{hoyer2024semivl} is an example that combines limited dense labels with a large unlabelled pool for semi-supervised learning. 

CLIPSurgery \cite{li2025CLIPSurgery} improves CLIP’s spatial localization by modifying its inference pathway to produce sharper similarity maps, which can be used to sample spatial prompts for SAM, and is a training-free method.

Unlike methods that treat CLIP either as the primary segmentation engine or solely as an external prompt generator, we use CLIP as a provider of semantic priors, whose features are injected directly into SAM’s encoder and jointly optimized with SAM.

\subsection{Integration of SAM and CLIP for Text-to-Mask Segmentation}

Existing VLM+SAM systems generally follow two strategies. The first relies on cascaded pipelines where a VLM converts text prompts into spatial prompts for SAM. Examples include GroundedSAM \cite{ren2024groundedSAM}, which uses GroundingDINO \cite{liu2024groundingdino} to generate box prompts, and CLIP-based methods that derive points or masks from similarity maps \cite{li2025CLIPSurgery, koleilat2409medclipsamv2, li2025clipsam, kollias2024sam2clip2sam}. Other approaches use SAM to propose masks, which are then filtered or scored using CLIP \cite{yang2024foundation, aleem2024salip}. In most cases, SAM and the VLM operate as separate stages.

A smaller body of work explores tighter feature-level coupling between SAM and VLMs by injecting language signals directly into SAM's internal representations \cite{wang2024sam, liu2025schnet, ma2025clisc, yu2024scnet, yuan2024open}. These approaches typically fuse text embeddings or activation maps through additional tokens, modified decoders, or auxiliary modules, often departing from the original SAM architecture or requiring substantial supervision and training, and are sometimes tailored to specific domains.

Our approach bridges these directions. We build on CLIPSurgery-style prompt strategy while also injecting CLIP-derived semantic features directly into SAM’s encoder via adapters, enabling internal semantic conditioning and joint SAM and CLIP co-adaptation with minimal architectural changes.

% Our approach is distinct from these frameworks: 
% we inject \emph{multi-modal} CLIP features—text, vision, and similarity—directly into SAM’s image encoder via lightweight parallel adapters, and \emph{jointly fine-tune} SAM and CLIP in both semi-automatic (text-only) and manual (text + points) modes. 
% By fusing semantics at the encoder level while preserving SAM’s prompt encoder and mask decoder, we obtain a parameter-efficient system that strengthens the coupling between CLIP and SAM rather than treating them as loosely connected stages.

\section{Method}

We propose \textbf{CLIP-Guided SAM}, which uses CLIP to produce (i) \emph{spatial semantic prompts} and (ii) \emph{feature-level conditioning prompts} that are injected into SAM’s image encoder via semantic adapters. We support two usage modes depending on whether point prompts come from CLIP (text-only inference) or from user/GT clicks (interactive setting).

\subsection{Generating Semantic Prompts with CLIP}

Given an image and a class prompt, CLIP produces:
(i) patch-level image embeddings 
$\mathbf{V} \in \mathbb{R}^{N \times C_c}$ 
(after discarding the [CLS] token, where $N = HW$), and 
(ii) a global text embedding 
$\mathbf{t} \in \mathbb{R}^{C_c}$.

We compute patch-wise cosine similarity:
\begin{equation}
\mathbf{s}_p =
\left\langle
\frac{\mathbf{V}_p}{\|\mathbf{V}_p\|},
\frac{\mathbf{t}}{\|\mathbf{t}\|}
\right\rangle,
\quad p=1,\dots,N,
\end{equation}
yielding similarity scores 
$\mathbf{s} \in \mathbb{R}^{N \times 1}$.

Reshaping $\mathbf{s}$ to $H \times W$ gives a similarity map $\mathbf{S}$.
We upsample and min–max normalize $\mathbf{S}$, then threshold it to obtain a binary prompt mask $\mathbf{B}$.

\begin{figure*}[t]
    \centering
    \includegraphics[width=\textwidth]{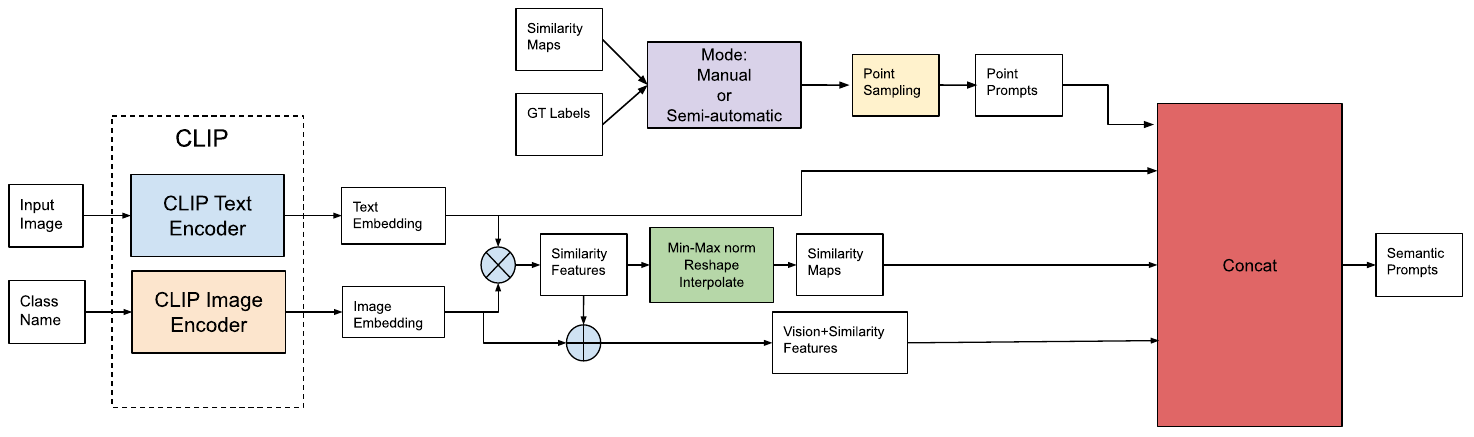}
    \vspace{-0.5em}
    \caption{
    \textbf{Semantic Prompt Generation with CLIP.} 
    Shows the process of how we use CLIP to generate our semantic prompts, which include similarity maps (dense mask prompt), vision+similarity features, text features, and point prompts. Point prompts are sampled either from similarity maps or GT labels, depending on the design mode (manual vs semi-automatic)
    }
    \vspace{-0.5em}
    \label{fig:semantic_prompts_from_clip}
\end{figure*}

We use these CLIP semantic signals in three ways (\cref{fig:semantic_prompts_from_clip}):

\begin{itemize}
    \item \textbf{Dense prompt:} resize $\mathbf{B}$ to SAM’s mask-prompt resolution and feed it to the prompt encoder.
    \item \textbf{Point prompts:} sample $K{=}5$ positive points from $\mathbf{B}$ (semi-automatic mode); in manual mode, points are sampled from GT masks.
    \item \textbf{Feature injection:} inject the text embedding $\mathbf{t}$ and fused vision–similarity features into SAM’s image encoder via semantic adapters.
\end{itemize}

\subsection{Two Design Modes}
\begin{figure*}[t]
    \centering
    \includegraphics[width=\textwidth]{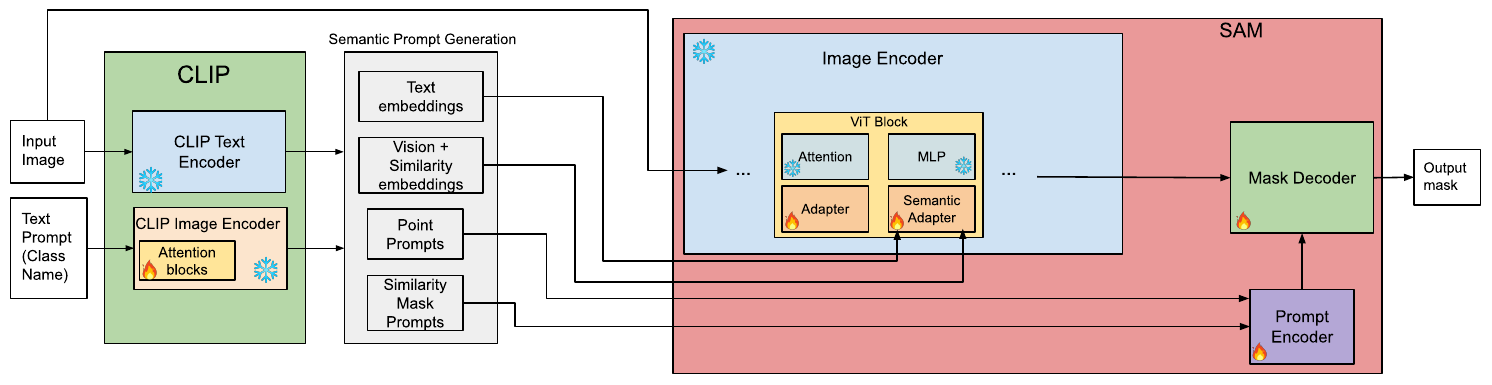}
    \vspace{-0.5em}
    \caption{
    \textbf{Overall integration framework.} 
    CLIP extracts text and image embeddings, which produce similarity features and maps via cosine similarity. 
    Points prompts are derived either from similarity masks or GT labels, depending on our design mode (manual or semi-automatic), while text, vision, and similarity features are injected into SAM through our semantic adapter modules.
    }
    \vspace{-0.5em}
    \label{fig:framework}
\end{figure*}

We provide two design modes in our integration framework (\cref{fig:framework}):

\textbf{Semi-automatic (text-only).} Points and dense prompts are derived from $\mathbf{B}$ during both training and inference, yielding train--test consistency under imperfect CLIP localization.

\textbf{Manual (interactive).} CLIP feature injection remains enabled and similarity maps are still used as dense prompts, but point prompts come from GT/user clicks (oracle) during both training and inference.

\subsection{Semantic Adapters}

\begin{figure}[t]
    \centering
    \includegraphics[width=\textwidth]{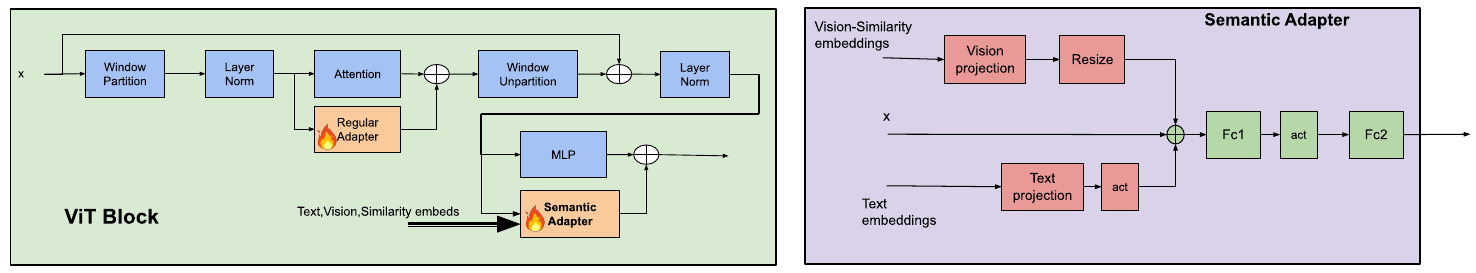}
    \vspace{-0.5em}
    \caption{
    \textbf{Adapter Placement and Internal Structure.}
    The figure on the left shows the structure of a ViT block in SAM's image encoder, and how our adapters are placed in parallel to the Attention and MLP blocks. The semantic adapter is in parallel to the MLP and receives the semantic inputs generated from CLIP.
    The right side shows the design of our semantic adapters. Inside each semantic adapter, text features are projected and fused through GELU, while vision–similarity (vis–sim) features are projected, spatially aligned, and added to SAM’s representation.
    }
    \label{fig:adapter}
    \vspace{-0.5em}
\end{figure}

We insert semantic adapters in parallel to the MLP of each ViT block (regular adapters remain parallel to attention). See~\cref{fig:adapter} for an overview of adapter placement and the internal structure of the semantic adapters.  

Let $\mathbf{F}_\ell \in \mathbb{R}^{H_\ell \times W_\ell \times C_\ell}$ denote the feature map entering the MLP at layer $\ell$. 
CLIP provides patch embeddings $\mathbf{V} \in \mathbb{R}^{N \times C_c}$, a text embedding $\mathbf{t} \in \mathbb{R}^{C_c}$, and similarity scores $\mathbf{s} \in \mathbb{R}^{N \times 1}$.  
We fuse vision and similarity via broadcasted addition,
\[
\mathbf{U} = \mathbf{V} + \mathbf{s},
\]
project to SAM’s channel dimension,
\[
\mathbf{U}' = \mathbf{U}\mathbf{W}_v \in \mathbb{R}^{N \times C_\ell},
\]
reshape to $H \times W \times C_\ell$, and interpolate to obtain
$\mathbf{U}_\ell \in \mathbb{R}^{H_\ell \times W_\ell \times C_\ell}$.
The text embedding is projected and activated,
\[
\mathbf{t}' = \mathrm{GELU}(\mathbf{W}_t \mathbf{t}) \in \mathbb{R}^{C_\ell},
\]
and broadcast spatially to $\mathbf{T}_\ell \in \mathbb{R}^{H_\ell \times W_\ell \times C_\ell}$.
Semantic conditioning is injected through residual addition:
\[
\hat{\mathbf{F}}_\ell =
\mathbf{F}_\ell +
\mathbf{U}_\ell +
\mathbf{T}_\ell,
\]
after which the adapter output is merged back into the main pathway by adding it to the output of the MLP block.

% \subsubsection{Feature Broadcasting Inside Semantic Adapters}

% We instantiate one semantic adapter in each transformer block of SAM’s image encoder; the adapters are not shared across layers. Each adapter is placed in parallel with the MLP block.

% Let $\mathbf{F}_\ell \in \mathbb{R}^{H_\ell \times W_\ell \times C}$ denote the feature map entering the MLP block at layer $\ell$. The CLIP text embedding is first projected to $\mathbb{R}^{C}$ and then broadcast across the spatial grid to form a constant conditioning map. This map is injected into the adapter pathway through element-wise addition.

% In parallel, the CLIP-derived vision and similarity features are projected and reshaped to $\mathbb{R}^{H_\ell \times W_\ell \times C}$ to match SAM’s internal feature resolution. This alignment is achieved through reshaping and interpolation. The resulting semantic feature map is fused with $\mathbf{F}_\ell$ inside the adapter branch via element-wise addition, before the adapter output is merged back into the transformer block.

% All semantic conditioning operations are performed within the semantic adapters, leaving the original transformer structure unchanged. Figure~\ref{fig:adapter} illustrates the internal design of the semantic adapter and its placement within each ViT block.

\subsection{Joint Optimization \& Training Objective}

We train SAM and CLIP end-to-end using supervision only on SAM's output masks. Since thresholding used for binary similarity mask prompt is non-differentiable, gradients reach CLIP only through the injected feature pathway.
We optimize a standard segmentation loss $\mathcal{L}=\mathcal{L}_{\mathrm{BCE}}+\mathcal{L}_{\mathrm{Dice}}+\mathcal{L}_{\mathrm{IoU}}$.
We fine-tune only CLIP's vision encoder's attention blocks while freezing the MLP blocks and the text encoder; for SAM, we fine-tune the prompt encoder, mask decoder, and the inserted adapters.

\subsection{Efficiency and Trainable Parameter Accounting}
\label{sec:efficiency}

Table~\ref{tab:param_accounting} reports total vs.\ trainable parameters. Overall, we train 49.0M parameters and deploy SAM+CLIP-vision (197.1M) under a closed-set assumption where class text embeddings are cached.  % (Move extra discussion to supp)
% ============================================================
% Table: Parameter Accounting
% ============================================================

\begin{table}[t]
\centering
\footnotesize
\setlength{\tabcolsep}{3pt}
\renewcommand{\arraystretch}{0.95}
\caption{\textbf{Parameter accounting of CLIP-Guided SAM (in millions).} 
``Trainable'' denotes parameters updated during training. 
``Deployment'' refers to the closed-set inference configuration where CLIP text embeddings are cached and only CLIP-vision is required at runtime.}
\label{tab:param_accounting}
\begin{tabular}{lcc}
\toprule
\textbf{Component} & \textbf{Total (M)} & \textbf{Trainable (M)} \\
\midrule
\textbf{SAM (overall)} & 110.3 & 20.6 \\
\quad Image encoder (incl.\ adapters) & 106.2 & 16.6 \\
\quad Mask decoder & 4.1 & 4.1 \\
\midrule
\textbf{CLIP (overall)} & 150.3 & 28.4 \\
\quad CLIP-vision & 86.8 & 28.3 \\
\quad CLIP-text & 63.4 & 0 \\
\midrule
\textbf{Full system total} & 260.5 & 49.0 \\
\textbf{Deployment total (SAM + CLIP-vision)} & 197.1 & 49.0 \\
\bottomrule
\end{tabular}

\vspace{0.35em}
\begin{minipage}{0.98\linewidth}
\footnotesize
\textit{Adapter overhead.} The SAM image-encoder adapters introduce 16.6M parameters, corresponding to an 18.5\% increase relative to the vanilla SAM backbone. 
In closed-set inference, CLIP text embeddings are precomputed once and the text encoder is not required at runtime.
\end{minipage}
\end{table}

\section{Experiments}

We evaluate on (i) camouflaged object detection (COD10K, CAMO, and CHAM-ELEON) and (ii) low-label multi-class segmentation (COCO, ADE20K, and Pascal VOC), with four experiments: SAM-PEFT baselines, VLM+SAM pipelines, comparison to SAM~3, and comparison to modern multi-class segmentation (semi-supervised and labelled\allowbreak-only).
Unless noted, we use SAM ViT-B, and use the CLIPSurgery~\cite{li2025CLIPSurgery} architecture for CLIP except in a few ablations; bold and underline mark best/second\allowbreak-best. We use the semi-automatic (text-only) mode in all experiments except Experiment 1, where we use both modes. For COCO, ADE20K, and Pascal VOC datasets, we use low-labelled-data splits commonly used in semi-supervised segmentation settings to simulate low-labelled-data scenarios~\cite{zou2020pseudoseg,hoyer2024semivl,yang2023unimatchv1,yang2025unimatchv2}.

\paragraph{Experiment 1: SAM PEFT baselines.}

We follow standard COD protocols on COD10K~\cite{Fan2022cod}, CAMO~\cite{le2019anabranchCAMO}, and CHAMELEON~\cite{skurowski2018animalchameleon}: train on COD10K+CAMO train splits; evaluate on their test splits and CHAMELEON; report MAE, $S_{\alpha}$, $E_{\phi}$, and $F_{\beta}^{w}$ following~\cite{Fan2022cod}. 
% Following~\cite{jalilian2025samptx} we use class names as prompts for COD10K (e.g., ``BatFish''); we use ``Camouflaged Object'' for the ``Other'' category and ``Camouflaged Animal'' for CAMO and CHAMELEON.

Table~\ref{tab:cod_results} shows that our manual and text-only variants achieve the best overall results, outperforming SAM-only PEFT baselines with and without manual prompts, supporting the value of active semantic supervision (both vision and text encoders of CLIP, and fine-tuning CLIP) and encoder-level semantic conditioning of SAM.

\begin{table}[t]
\centering
\footnotesize
\setlength{\tabcolsep}{1.5pt}
\renewcommand{\arraystretch}{0.7}
\caption{First-experiment results on camouflaged object datasets: CHAMELEON, CAMO, and COD10K.}
\vspace{0.25em}
\resizebox{\linewidth}{!}{%
\begin{tabular}{l|cccc|cccc|cccc}
\toprule
\textbf{Method} &
\multicolumn{4}{c|}{\textbf{CHAMELEON}} &
\multicolumn{4}{c|}{\textbf{CAMO}} &
\multicolumn{4}{c}{\textbf{COD10K}} \\
& \makecell{$S_{\alpha}$ \\ (↑)} & \makecell{$E_{\phi}$ \\ (↑)} & \makecell{$F_{\beta}^{w}$ \\ (↑)} & \makecell{MAE \\ (↓)} 
& \makecell{$S_{\alpha}$ \\ (↑)} & \makecell{$E_{\phi}$ \\ (↑)} & \makecell{$F_{\beta}^{w}$ \\ (↑)} & \makecell{MAE \\ (↓)} 
& \makecell{$S_{\alpha}$ \\ (↑)} & \makecell{$E_{\phi}$ \\ (↑)} & \makecell{$F_{\beta}^{w}$ \\ (↑)} & \makecell{MAE \\ (↓)} \\
\midrule
\multicolumn{13}{l}{\emph{(A) Methods requiring manual point prompts}} \\
SAM~\cite{kirillov2023segment}                                   & 0.727 & 0.734 & 0.639 & 0.081 & 0.684 & 0.687 & 0.606 & 0.132 & 0.783 & 0.798 & 0.701 & 0.050 \\
\makecell[l]{SU-SAM-Series~\cite{song2024SUSAM}} & 0.762 & 0.824 & 0.552 & 0.083 & 0.671 & 0.779 & 0.489 & 0.139 & 0.720 & 0.729 & 0.455 & 0.073 \\
\makecell[l]{SU-SAM-Mix~\cite{song2024SUSAM}}    & 0.915 & 0.918 & 0.819 & 0.029 & 0.861 & 0.895 & 0.763 & 0.065 & 0.881 & 0.868 & 0.751 & 0.029 \\
\makecell[l]{SU-SAM-Parallel~\cite{song2024SUSAM}} & \textbf{0.930} & \underline{0.940} & \underline{0.856} & \underline{0.024} & {0.890} & \underline{0.925} & \underline{0.815} & \underline{0.053} & \textbf{0.915} & \underline{0.909} & \underline{0.818} & \underline{0.020} \\
\makecell[l]{SU-SAM-LoRA~\cite{song2024SUSAM}}   & 0.689 & 0.764 & 0.384 & 0.130 & 0.641 & 0.747 & 0.379 & 0.168 & 0.671 & 0.669 & 0.324 & 0.097 \\
\makecell[l]{SAM-PTx~\cite{jalilian2025samptx}}                 & \textbf{0.930} & \underline{0.940} & {0.853} & \underline{0.024} & \underline{0.891} & 0.921 & 0.810 & \underline{0.053} & \textbf{0.915} & 0.906 & 0.813 & \underline{0.020} \\
Ours (manual)                         & \underline{0.927} & \textbf{0.967} & \textbf{0.903} & \textbf{0.017} & \textbf{0.896} & \textbf{0.944} & \textbf{0.855} & \textbf{0.035} & \underline{0.914} & \textbf{0.965} & \textbf{0.874} & \textbf{0.015} \\
\midrule
\multicolumn{13}{l}{\emph{(B) Methods without manual point prompts}} \\
\textbf{†}SINetV2~\cite{Fan2022cod}                     & 0.888 & 0.942 & 0.816 & 0.030 & 0.820 & 0.882 & 0.743 & \underline{0.070} & 0.815 & 0.887 & 0.680 & 0.037 \\
\makecell[l]{SAM-Adapter~\cite{chen2023SamAdptr}}             & 0.896 & 0.919 & 0.824 & 0.033 & \underline{0.847} & 0.873 & \underline{0.765} & \underline{0.070} & \underline{0.883} & \underline{0.918} & \underline{0.801} & \underline{0.025} \\
\makecell[l]{Conv-LoRA~\cite{zhong2024convMLoRA}}              & 0.891 & 0.922 &  0.790 & 0.031 & 0.832 & \underline{0.865} & 0.741 & 0.073 & 0.861 & 0.903  & 0.747 & 0.031 \\
\makecell[l]{TS-SAM~\cite{yu2024tssam}}                  & \underline{0.912} & \underline{0.947} & \underline{0.849} & \underline{0.023} & 0.826 & 0.862 & 0.753 & 0.073 & 0.863 & 0.909 & 0.771 & 0.029 \\
Ours (text-only)                 & \textbf{0.923} & \textbf{0.970} & \textbf{0.905} & \textbf{0.018} & \textbf{0.877} & \textbf{0.924} & \textbf{0.845} & \textbf{0.044} & \textbf{0.899} & \textbf{0.951} & \textbf{0.854} & \textbf{0.018} \\
\bottomrule
\end{tabular}}%
\vspace{0.35em}
\begin{minipage}{\textwidth}
\footnotesize{\textit{†\,Note:} SINetV2 is not SAM-based but is included as a key baseline for camouflaged object detection, as it was introduced by the authors of the COD10K dataset and remains a strong benchmark reference.}
\vspace{-0.4em}
\end{minipage}
\label{tab:cod_results}
\end{table}

\paragraph{Experiment 2: VLM+SAM baselines.}
We compare against representative vision–language SAM pipelines, including GroundedSAM~\cite{ren2024groundedSAM} and CLIPSurgery~\cite{li2025CLIPSurgery} with SAM, across COCO, ADE20K, PASCAL VOC, and camouflaged object benchmarks. GroundedSAM performs external grounding via GroundingDINO \allowbreak~\cite{liu2024groundingdino}, while CLIPSurgery uses CLIP-based similarity maps to generate prompts for SAM.

For CLIPSurgery, we report the original zero-shot model as well as progressively stronger variants that add dense mask prompts from similarity maps, decoder fine-tuning, and adapters. For COCO/ADE20K/VOC, our model is trained on the split with the least supervision.

Table~\ref{tab:second_exp} shows that our jointly trained framework consistently outperforms CLIPSurgery variants across all datasets and GroundedSAM on all benchmarks except COCO (by a small margin). Notably, even when CLIPSurgery is strengthened with decoder fine-tuning and adapters, our method achieves substantially better performance despite using the same CLIP backbone, indicating that the gains arise from internal semantic conditioning and joint SAM–CLIP co-adaptation rather than from model size or training alone.

The largest improvements appear on camouflaged object benchmarks, where purely spatial grouping is insufficient. While fine-tuning improves CLIPSurgery variants, external semantic grounding alone remains insufficient; jointly adapting SAM and CLIP through internal semantic conditioning proves critical for robust segmentation in these challenging scenarios.

\begin{table}[t]
\centering
\footnotesize
\setlength{\tabcolsep}{1.5pt}
\renewcommand{\arraystretch}{0.7}
\caption{Second-experiment results across datasets. Class mIoU is reported for PASCAL, ADE20K, and COCO; MAE is reported for COD10K, CAMO, and CHAMELEON.}
\label{tab:second_exp}
{
\resizebox{\textwidth}{!}{%
\begin{tabular}{l c l c c c c c c}
\toprule
\textbf{Method} & \makecell{\textbf{Fine-tuning}\\\textbf{SAM?}} & \textbf{VLM Size} &
\makecell{\textbf{PASCAL}\\(mIoU \%)} &
\makecell{\textbf{ADE20K}\\(mIoU \%)} &
\makecell{\textbf{COCO}\\(mIoU \%)} &
\makecell{\textbf{COD10K}\\(MAE)} &
\makecell{\textbf{CAMO}\\(MAE)} &
\makecell{\textbf{CHAMELEON}\\(MAE)} \\
\midrule
\makecell[l]{GroundedSAM~\cite{ren2024groundedSAM}} & No  & \makecell[l]{172M (GroundingDINO)} & \underline{72.0} & \underline{42.4} & \textbf{61.3} & 0.500 & 0.146 & 0.249 \\
\makecell[l]{CLIPSurgery+SAM (vanilla)~\cite{li2025CLIPSurgery}}                                         & No  & \makecell[l]{87M (CS-CLIP-Base)}     & 41.2 & 20.0   & 23.9    & 0.533    & 0.478    & 0.253    \\
\makecell[l]{\textbf{†}CLIPSurgery+SAM (modified)~\cite{li2025CLIPSurgery}}                                          & No & \makecell[l]{87M (CS-CLIP-Base)}     & 48.6 & 23.4 & 28.0 & 0.287 & 0.527 & 0.501 \\

\makecell[l]{CLIPSurgery+SAM~\cite{li2025CLIPSurgery}}                                         & Yes (decoder)  & \makecell[l]{87M (CS-CLIP-Base)}     & 58.8 & 23.1  & 34.6  & 0.045  & 0.100   & 0.046   \\
\makecell[l]{CLIPSurgery+SAM~\cite{li2025CLIPSurgery}}                                         & Yes (adapters + decoder)  & \makecell[l]{87M (CS-CLIP-Base)}     & 61.4 & 24.8  & 36.2   & \underline{0.026} & \underline{0.063}   & \underline{0.020}   \\
\makecell[l]{CLIP-Guided SAM (ours)} & Yes & \makecell[l]{87M (CS-CLIP-Base)}     & \textbf{78.5} & \textbf{47.9} & \underline{60.5} & \textbf{0.018} & \textbf{0.044} & \textbf{0.018} \\
\bottomrule
\end{tabular}%
}}
\begin{minipage}{\textwidth}
\tiny{\textit{†\,Note:} the modified, training-free CLIPSurgery+SAM baseline includes binary similarity maps as mask prompts in addition to points prompts.} \\
\tiny{The fine-tuned results use the dataset with the least amount of data for COCO/ADE/PASCAL, which are 1/512, 1/64, and 1/16 respectively.}
\end{minipage}
\end{table}

\paragraph{Experiment 3: Direct Comparison with SAM 3.}

SAM~3~\cite{carion2025sam3segmentconcepts} is a large-scale segmentation foundation model with built-in text-prompt capability trained on massive datasets. Unlike SAM~3, which is trained end-to-end for concept segmentation on massive datasets, our method relies on parameter-efficient semantic injection and co-adaptation using limited supervision.

We evaluate SAM~3 both zero-shot and with light fine-tuning on COCO 1/512, training only its segmentation head and transformer-based fusion components (23M parameters). Even under this limited fine-tuning regime, SAM~3 achieves strong performance. For fair comparison, predictions are converted to per-class binary masks and evaluated using the same class-wise mIoU protocol as our method.

Table~\ref{tab:sam3_exp} shows that with limited supervision, our smaller system (197M deployment parameters) approaches or surpasses SAM~3 zero-shot performance and remains competitive as supervision increases.

To examine scaling behaviour, we also evaluate two larger backbone variants: 
CLIP(ViT-L/14)+SAM(ViT-B) and CLIP(ViT-B/16)+SAM(ViT-H). 
Increasing CLIP capacity yields a substantial improvement (65.0 mIoU on COCO 1/512), whereas increasing SAM capacity alone provides a smaller gain (61.2 mIoU). 
This suggests that strengthening the semantic guide is more beneficial than scaling the segmentation backbone within our framework.

\begin{table}[t]
\centering
\caption{Direct comparison with SAM 3~\cite{carion2025sam3segmentconcepts} on COCO. Results are mIoU.}
\label{tab:sam3_exp}
\setlength{\tabcolsep}{6pt}
{\scriptsize
\renewcommand{\arraystretch}{1.05}
\resizebox{\textwidth}{!}{%
\begin{tabular}{l c c c c c c c c}
\toprule
\textbf{Method} &
\makecell{\textbf{Total}\\\textbf{Params}} &
\makecell{\textbf{Trainable}\\\textbf{Params}} &
\textbf{Zero-shot} &
\makecell{\textbf{1/512}\\(232)} &
\makecell{\textbf{1/256}\\(463)} &
\makecell{\textbf{1/128}\\(925)} &
\makecell{\textbf{1/64}\\(1849)} &
\makecell{\textbf{1/32}\\(3697)} \\
\midrule
SAM 3 & 840M & 23M* & 63.7 & 68.8 & -- & -- & -- & -- \\
\midrule
Ours (CLIP-B + SAM-B) & 197M$^\dagger$ & 49M & 28.0 & 60.5 & 63.2 & 65.8 & 67.7 & 69.2 \\
Ours (CLIP-L + SAM-B) & 543M & 126M & -- & 65.0 & -- & -- & -- & -- \\
Ours (CLIP-B + SAM-H) & 809M & 50M & -- & 61.2 & -- & -- & -- & -- \\
\bottomrule
\end{tabular}
}}

\vspace{0.5em}
\footnotesize{
*SAM 3 fine-tuning trains the segmentation head and transformer layers only.  
$^\dagger$197M corresponds to SAM-B + added adapters + CLIP vision backbone at inference. Text embeddings can be precomputed under closed-set assumptions.
}
\end{table}

\paragraph{Experiment 4: Comparison to multi-class segmentation.}
This comparison is not task-identical: prior methods predict all classes jointly, whereas we predict a binary mask per class prompt and assume image-level class presence via text prompts. Nevertheless, it provides a useful reference point for the extent to which prompt-conditioned segmentation can go when class intent is available.

Table~\ref{tab:third_exp_combined} shows strong gains on COCO and ADE20K, despite our method using no unlabelled data while several baselines rely on large semi-supervised training sets (SemiVL~\cite{hoyer2024semivl}, UniMatchV2~\cite{yang2025unimatchv2}). Compared to the labelled-only variant of UniMatchV2, our results highlight the effectiveness of prompt-conditioned segmentation even without additional data. Importantly, these improvements are not explained by backbone capacity alone: a CLIP+SAM baseline with similar backbones but without semantic adapters performs substantially worse (see Experiment~2 and Table~\ref{tab:second_exp}).

Performance differences are smaller on PASCAL VOC, a simpler dataset where prompt-based conditioning offers less advantage and factors such as backbone capacity and semi-supervised training become more influential.

\begin{table*}[t]
\centering
\caption{\textbf{Experiment 4: Comparison to multi-class segmentation.}
Class mIoU (\%) on COCO, ADE20K, and PASCAL VOC under varying label fractions. 
Baselines perform joint multi-class prediction, whereas our method predicts a binary mask per class prompt using labelled data only (no unlabelled data).}\label{tab:third_exp_combined}
\scriptsize
\setlength{\tabcolsep}{2.6pt}
\renewcommand{\arraystretch}{1.05}
\resizebox{\textwidth}{!}{%
\begin{tabular}{l c c l l | c c c c c | c c c c c | c c c c c}
\toprule
\textbf{Method} &
\makecell{\textbf{Unlab}\\\textbf{data?}} &
\makecell{\textbf{Img-level}\\\textbf{prompts?}} &
\textbf{Encoder} &
\makecell{\textbf{\# Params}} &
\multicolumn{5}{c|}{\textbf{COCO (mIoU \%)}} &
\multicolumn{5}{c|}{\textbf{ADE20K (mIoU \%)}} &
\multicolumn{5}{c}{\textbf{PASCAL VOC (mIoU \%)}} \\
\cmidrule(lr){6-10}
\cmidrule(lr){11-15}
\cmidrule(lr){16-20}
& & & & &
\makecell{\textbf{1/512}\\\tiny (232)} &
\makecell{\textbf{1/256}\\\tiny (463)} &
\makecell{\textbf{1/128}\\\tiny (925)} &
\makecell{\textbf{1/64}\\\tiny (1849)} &
\makecell{\textbf{1/32}\\\tiny (3697)} &
\makecell{\textbf{1/64}\\\tiny (316)} &
\makecell{\textbf{1/32}\\\tiny (631)} &
\makecell{\textbf{1/16}\\\tiny (1262)} &
\makecell{\textbf{1/8}\\\tiny (2526)} &
\makecell{\textbf{1/4}\\\tiny (5052)} &
\makecell{\textbf{1/16}\\\tiny (92)} &
\makecell{\textbf{1/8}\\\tiny (183)} &
\makecell{\textbf{1/4}\\\tiny (366)} &
\makecell{\textbf{1/2}\\\tiny (732)} &
\makecell{\textbf{Full}\\\tiny (1464)} \\
\midrule
SemiVL~\cite{hoyer2024semivl} &
Yes & No & CLIP-Base & 88M &
50.1 & 52.8 & 53.6 & 55.4 & 56.5 &
33.7 & 35.1 & 37.2 & 39.4 & -- &
84.0 & 85.6 & 86.0 & 86.7 & 87.3 \\
UniMatchV2~\cite{yang2025unimatchv2} &
Yes & No & DINOv2-B & 97.5M &
47.9 & 55.8 & 58.7 & 60.4 & 63.3 &
38.7 & 45.0 & 46.7 & 49.8 & 52.0 &
\textbf{86.3} & \textbf{87.9} & \textbf{88.9} & \textbf{90.0} & \textbf{90.8} \\
UniMatchV2~\cite{yang2025unimatchv2} (labelled-only) &
No & No & DINOv2-B & 97.5M &
36.8 & 45.8 & 52.1 & 56.2 & 59.5 &
26.1 & 39.3 & 42.8 & 46.4 & 49.0 &
76.9 & 82.1 & 85.3 & 87.2 & 88.3 \\
Our Method (labelled-only) &
No & Yes & \makecell[l]{CLIP-B+SAM-B} &
\makecell[l]{197M} &
\textbf{60.5} & \textbf{63.2} & \textbf{65.8} & \textbf{67.7} & \textbf{69.2} &
\textbf{47.9} & \textbf{51.6} & \textbf{55.2} & \textbf{57.7} & \textbf{61.1} &
78.3 & 81.7 & 82.9 & 83.9 & 85.0 \\
\bottomrule
\end{tabular}%
}
\vspace{0.3em}
\end{table*}

% ============================================================
% Efficiency / Parameter Accounting Section (Add-on 1)
% Place after Experiment 3 (recommended) or at end of Experiment 1
% ============================================================

\subsection{Ablation Studies}
\label{sec:ablation}
% We conduct ablation experiments on the PASCAL VOC 1/16 split in the semi-automatic (text-only) setting and report validation mIoU to assess the contribution of each component. We selected this extremely low-labelled split (92 labelled images) for efficiency in running the experiments. However, the extremely small number of labels could affect the results. For example, small differences in results might become more significant if trained with more data. There may also be overtraining in some cases, which could complicate the analysis. That said, the general and important trends and patterns observed are sufficiently robust to be unaffected by these potential issues.
Unless stated otherwise, ablations use PASCAL VOC 1/16 in the semi-automatic setting and are evaluated using mIoU.

% \paragraph{Multi-modal feature injection.}
% Table~\ref{tab:ablation_modalities} shows that full fusion yields the best result (78.3 mIoU). Similarity features dominate: any setup retaining them stays high (77.5–78.3), whereas removing similarity drops performance to the 66–69 range. Text-only, vision-only, and text+vision remain weak without similarity.
% At the bottom, no adapters gives the lowest score (66.6), followed by parallel adapters without any modality input (67.1), indicating that structural adapters alone offer minimal benefit. The consistent gains from adding text, vision, or—especially—similarity confirm that injecting multi-modal cues is essential for strong performance.

\paragraph{Which injected modality matters?}
Table~\ref{tab:ablation_modalities} shows that similarity features provide the dominant signal: removing similarity causes a sharp performance drop, while similarity-only remains strong (77.5 mIoU) and almost matches the full method. In contrast, text-only and vision-only variants perform substantially worse. This means that similarity features carry the main semantic localization, and text and vision features refine it slightly. Furthermore, adapters without semantic inputs provide minimal benefit (67.1 vs.\ 66.6 without adapters), confirming that semantic injection—rather than having regular adapters alone—is the main source of improvement. 

\begin{table}[t]
\centering
\scriptsize
\setlength{\tabcolsep}{20pt}
\renewcommand{\arraystretch}{1.03}
\caption{\textbf{Ablation on feature modalities injected into adapters (VOC 1/16).}}
\label{tab:ablation_modalities}
\resizebox{\textwidth}{!}{%
\begin{tabular}{c c c c c c}
\toprule
\textbf{Text} & \textbf{Vision} & \textbf{Sim.} & \textbf{Adapt.} & \textbf{Config} & \textbf{mIoU} \\
\midrule
\cmark & \cmark & \cmark & \cmark & Full fusion & \textbf{78.3} \\
\midrule
\cmark & \cmark & \xmark & \cmark & Text+Vision & 69.5 \\
\cmark & \xmark & \cmark & \cmark & Text+Similarity & {78.1} \\
\xmark & \cmark & \cmark & \cmark & Vision+Similarity & 77.5 \\
\midrule
\cmark & \xmark & \xmark & \cmark & Text only & 67.2 \\
\xmark & \cmark & \xmark & \cmark & Vision only & 68.8 \\
\xmark & \xmark & \cmark & \cmark & Sim. only & 77.5 \\
\midrule
\xmark & \xmark & \xmark & \cmark & Parallel only & 67.1 \\
\xmark & \xmark & \xmark & \xmark & No adapters & 66.6 \\
\bottomrule
\end{tabular}}
\vspace{-1pt}
\end{table}

\paragraph{Co-adaptation budget.}
We vary $(C,S)$: the number of trainable CLIP vision blocks (attention) and the number of SAM blocks equipped with semantic \allowbreak adapters. 
Table~\ref{tab:coadapt_budget} reveals an asymmetric effect: performance is more sensitive to CLIP adaptation than to increasing SAM semantic depth. 
While increasing the number of trainable CLIP blocks consistently improves or stabilizes performance, expanding semantic adapters across SAM layers yields only marginal gains, suggesting that SAM requires only modest semantic capacity once strong semantic guidance is available. 
This trend is consistent with the scaling results in Experiment~3, where increasing CLIP capacity (CLIP-L) provided larger gains than scaling the SAM backbone (SAM-H), indicating that the semantic encoder plays a dominant role in the framework’s performance.

% ============================================================
% Co-Adaptation Budget Table (expanded)
% ============================================================

\begin{table}[t]
\centering
\footnotesize
\setlength{\tabcolsep}{4pt}
\renewcommand{\arraystretch}{0.95}
\caption{\textbf{Co-adaptation budget study on PASCAL VOC 1/16 (mIoU \%).}
$(C,S)$ denotes the number of CLIP vision blocks (attention weights fine-tuned) and SAM blocks equipped with trainable semantic adapters.}
\label{tab:coadapt_budget}
\begin{tabular}{lccc}
\toprule
\textbf{SAM/CLIP} & \textbf{$C{=}4$} & \textbf{$C{=}8$} & \textbf{$C{=}12$} \\
\midrule
$S{=}4$  & 77.4 & 77.2 & {78.2} \\
$S{=}8$  & 77.0 & 77.6 & 78.0 \\
$S{=}12$ & 77.1 & 77.17 & 78.1 \\
\bottomrule
\end{tabular}
\end{table}

% \paragraph{Dense similarity prompting.}
% Disabling dense similarity prompting (using similarity maps only for point sampling) reduces performance from 78.3 to 77.6 mIoU, confirming that dense similarity masks provide useful spatial conditioning beyond discrete points.

\paragraph{Train-test prompt alignment and dense similarity prompting.}
Using similarity maps as dense mask prompts provides additional spatial conditioning beyond point prompts (78.3$\rightarrow$77.6). More importantly, training with CLIP-sampled points significantly improves robustness: models trained on GT points but evaluated on CLIP prompts drop sharply (78.3$\rightarrow$69.8), supporting our strategy of training with noisy prompts for text-only inference.

\paragraph{Analyzing CLIP adaptation.}
We analyze how the choice and training strategy of the semantic guide (CLIP) affects performance. 
Replacing CLIPSurgery with standard CLIP reduces accuracy substantially (78.3$\rightarrow$72.8 mIoU), confirming that spatially aligned similarity maps are critical for effective conditioning. 
Freezing CLIPSurgery also degrades performance (73.7 mIoU), indicating that jointly optimizing the semantic guide and the segmentation backbone is important.

Interestingly, pre-fine-tuning CLIPSurgery on the target dataset improves its standalone mask quality but leads to worse final performance after joint training (Table~\ref{tab:clip_prefinetune_summary}). 
Although the pre-adapted model starts from a stronger initialization, it converges to a substantially lower final accuracy (67.5 mIoU vs.\ 78.3), suggesting that strong prior specialization can hinder SAM--CLIP co-adaptation and may even underperform weaker semantic guides such as standard CLIP (72.8) when integrated into the joint system. 
This suggests that maximizing the standalone performance of a vision–language model does not necessarily translate to better segmentation performance when the model is later integrated into a jointly trained system.

Overall, these results indicate that effective semantic conditioning emerges from joint optimization rather than from independently maximizing the quality of the semantic encoder.

\begin{table}[t]
\centering
\scriptsize
\setlength{\tabcolsep}{20pt}
{\renewcommand{\arraystretch}{1.05}
\vspace{-1pt}
\caption{\textbf{Effect of pre-finetuning CLIP on our CLIP+SAM framework (VOC 1/16).} Values are mIoU.}
\label{tab:clip_prefinetune_summary}
\resizebox{\textwidth}{!}{%
\begin{tabular}{lcccc}
\toprule
\textbf{Setup} &
\textbf{Stage} &
\textbf{Start} &
\textbf{Final} &
$\Delta$ \\
\midrule
CLIP pre-finetune &
100 epochs & -- & 58 & -- \\
\midrule
SAM+CLIP (no pre) &
Joint & 38.7 & 78.3 & +35.7 \\
SAM+CLIP (with pre) &
Joint & 54.1 & 67.5 & +12.4--14.4 \\
\bottomrule
\end{tabular}}}
\vspace{-1pt}
\end{table}

% \paragraph{Loss function.}
% Table~\ref{tab:loss_ablation} compares several BCE/Dice/IoU combinations on VOC 1/16. BCE-only gives the lowest score (76.1 mIoU), while adding a second term consistently helps, with BCE+IoU (78.1) outperforming BCE+Dice and Dice+IoU. The best result is obtained when all three terms are weighted equally (78.3).

\paragraph{Loss.}
Table~\ref{tab:loss_ablation} shows that combining BCE, Dice, and IoU yields the best performance.

\begin{table}[t]
\centering
\scriptsize
\setlength{\tabcolsep}{20pt}
\renewcommand{\arraystretch}{1.05}
\vspace{-1pt}
\caption{\textbf{Ablation on loss function (VOC 1/16).} Values are mIoU (\%).}
\label{tab:loss_ablation}
\begin{tabular}{l c}
\toprule
\textbf{Loss} & \textbf{mIoU (\%)} \\
\midrule
BCE & 76.1 \\
BCE + Dice & 77.7 \\
BCE + IoU & 78.1 \\
Dice + IoU & 76.8 \\
BCE + Dice + IoU & \textbf{78.3} \\
\bottomrule
\end{tabular}
\vspace{-1pt}
\end{table}

% \subsection{Qualitative Analysis}
% \label{sec:qualitative}

% Figure~\ref{fig:ade16_qualitative} provides qualitative comparisons on ADE20K (1/16 split). 
% Each row shows the input image, ground-truth mask, and outputs from several configurations: 
% Vanilla SAM with manual points, Vanilla SAM with CLIP- and CLIPSurgery-sampled points, 
% our CLIP-Guided SAM with CLIPSurgery prompts but no fine-tuning, 
% and our jointly fine-tuned CLIP-Guided SAM. 
% The fully fine-tuned model produces more complete and semantically consistent segmentations, recovering fine object boundaries and details that baseline variants miss, and approaches the quality of Vanilla SAM with accurate GT-sourced point prompts in semi-automatic mode without manual spatial clicks. 
% These visual results support our quantitative findings that multi-modal CLIP guidance and joint SAM+CLIP fine-tuning improve robustness to noisy or sparse prompts.

\subsection{Qualitative Analysis}
\label{sec:qualitative}
\begin{figure}[t]
    \centering
    \includegraphics[width=\textwidth]{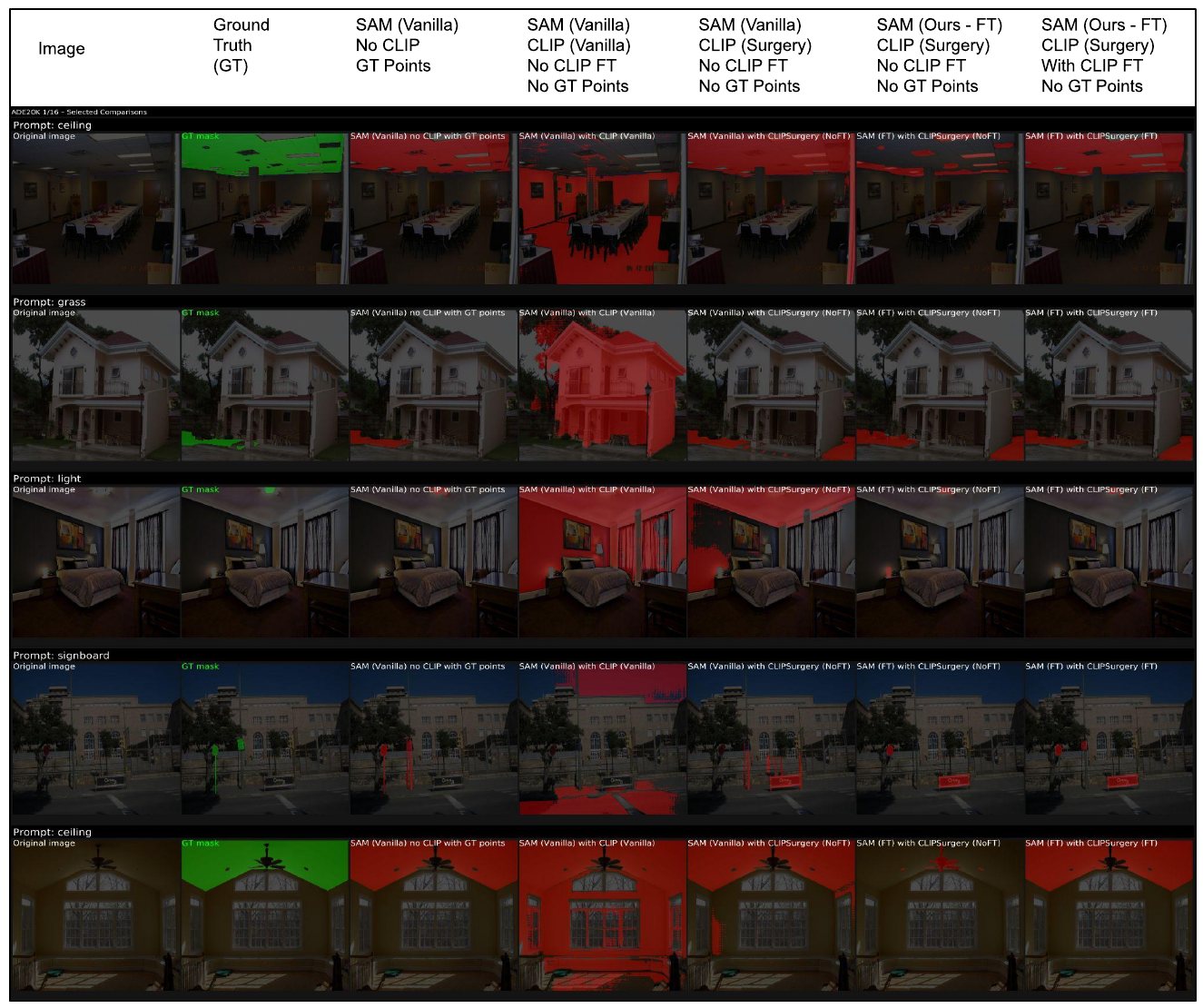}
    \vspace{-0.5em}
    \caption{
    \textbf{Qualitative analysis on ADE20K (1/16 split).} Comparison of vanilla SAM, CLIP-based prompting variants, and our CLIP-Guided SAM before and after fine-tuning.}
    % Visual comparison of five representative examples. 
    % Each row shows, from left to right: 
    % (1) input image, (2) ground-truth mask, 
    % (3) Vanilla SAM with manual point prompts, 
    % (4) Vanilla SAM with CLIP-sampled points, 
    % (5) Vanilla SAM with CLIPSurgery-sampled points~\cite{li2025CLIPSurgery}, 
    % (6) our \emph{CLIP-Guided SAM} using CLIPSurgery-sampled points (no fine-tuning), and 
    % (7) our \emph{CLIP-Guided SAM} jointly fine-tuned with CLIP on ADE20K (1/16). 
    % Our fine-tuned model produces more complete and semantically aligned masks, particularly in cluttered or ambiguous regions.
    % }
    \vspace{-0.5em}
    \label{fig:ade16_qualitative}
\end{figure}

Figure~\ref{fig:ade16_qualitative} shows qualitative comparisons on ADE20K (1/16 split) across vanilla SAM, CLIP- and CLIPSurgery-driven variants, and our CLIP-Guided SAM with and without fine-tuning. The jointly fine-tuned model produces more complete and semantically consistent masks, recovering fine object boundaries and cluttered regions where baselines either miss objects or leak into distractors, corroborating the quantitative gains of our multi-modal encoder injection and joint SAM-CLIP training.

\section{Conclusion}
\label{sec:conclusion}

We introduced \textit{CLIP-Guided SAM}, a parameter-efficient framework for enabling \emph{semantic conditioning} in the Segment Anything Model under limited supervision and compute. 
Instead of treating a vision--language model as an external prompt generator, our approach injects CLIP-derived semantic signals directly into SAM’s image encoder through semantic adapters, allowing semantics to influence feature formation while preserving SAM’s promptable design. 
The same mechanism supports both interactive segmentation with user-provided points and a semi-automatic text-only regime, in which spatial prompts are derived from CLIP similarity maps.

Our experiments show that effective semantic conditioning requires \emph{joint co-adaptation} between the segmentation backbone and its semantic guide. 
Freezing or independently pre-training CLIP degrades performance, while end-to-end co-adaptation enables CLIP to serve as a task-aligned semantic guide for SAM. 
Across camouflaged object benchmarks and low-label segmentation datasets, including COCO, ADE20K, and Pascal VOC, CLIP-Guided SAM consistently outperforms SAM-based PEFT methods and VLM+SAM pipelines while remaining competitive with powerful promptable segmentation foundation models such as SAM~3 under limited compute and supervision.

Overall, these results suggest that internal semantic co-adaptation provides a practical alternative to shallow prompt-based coupling when adapting promptable segmentation models under limited data and compute.

% ---------------------------------------------------------------
% Acknowledgements (REMOVE in review; add for camera-ready)
% \section*{Acknowledgements}
% This work was supported by ...

% ---------------------------------------------------------------
% Bibliography (ECCV/LNCS style)
% IMPORTANT:
% - ECCV uses BibTeX with splncs04, NOT biblatex/\printbibliography.
% - Point this to your .bib file (e.g., main.bib / references.bib).
\bibliographystyle{splncs04}
\bibliography{main}

\end{document}